\documentclass[sigplan,10pt,nonacm]{acmart}
\usepackage{algorithm}
\usepackage{algorithmic}
\usepackage{xspace}

\newcommand{\utf}[1]{{\sc{utf}}{\tt{}-}{\small\tt{}#1}{}\xspace}
\newcommand{\ascii}{{\sc{}ascii}{}\xspace}
\newcommand{\unicode}{{\sc{}u}nicode{}\xspace}

\newcommand{\typz}{$\mathbb{Z}$}

\newcommand{\java}{\texttt{Java}{}\xspace}
\newcommand{\cpp}{\texttt{C++}{}\xspace}
\newcommand{\rust}{\texttt{Rust}{}\xspace}

\newlength\myindent
\newcommand{\mafigpdf}[3]{%
\begin{figure}[ht]%
\begin{center}%
\includegraphics[scale=#1]{#2.pdf}%
\end{center}%
\vspace{-10pt}%
\Description{#3}%
\caption{#3}%
\label{fig-#2}%
\end{figure}%
}

\usepackage[english]{babel}

\AtBeginDocument{%
  }

\setcopyright{acmcopyright}
\copyrightyear{2023}
\acmYear{2023}
\acmDOI{XXXXXXX.XXXXXXX}

\acmConference[Submission to PLOS'23]{}{October 23, 2023}{Koblenz, Germany}
\acmPrice{15.00}
\acmISBN{978-1-4503-XXXX-X/18/06}

\begin{document}

\title[Tips for making the most of 64-bit architectures]{Tips for making the most of 64-bit architectures in langage design, libraries or garbage collection}

\author{Benoît Sonntag}
\authornote{Both authors contributed equally to this research.}
\email{Benoit.Sonntag@lisaac.org}
\affiliation{%
  \institution{Université de Strasbourg}
  \city{Strasbourg}
  \country{France}
}

\author{Dominique Colnet}
\email{Dominique.Colnet@loria.fr}
\affiliation{%
  \institution{Université de Lorraine}
  \city{Nancy}
  \country{France}}

\begin{abstract}
\noindent
The 64-bit architectures that have become standard today offer unprecedented low-level programming possibilities.  
For the first time in the history of computing, the size of address registers far exceeded the physical capacity of their bus.
After a brief reminder of the possibilities offered by the small size of addresses compared to the available 64 bits,
we develop three concrete examples of how the vacant bits of these registers can be used.
Among these examples, two of them concern the implementation of a library for a new statically typed programming language.
Firstly, the implementation of multi-precision integers, with the aim of improving performance in terms of both calculation speed and RAM savings.
The second example focuses on the library's handling of \utf{8} character strings.
Here, the idea is to make indexing easier by ignoring the physical size of each \utf{8} characters.
Finally, the third example is a possible enhancement of garbage collectors, in particular the mark \& sweep for the object marking phase.
\end{abstract}

%%
%% The code below is generated by the tool at http://dl.acm.org/ccs.cfm.
%% Please copy and paste the code instead of the example below.
%%

%%
%% Keywords. The author(s) should pick words that accurately describe
%% the work being presented. Separate the keywords with commas.
\keywords{64 bits architecture, address bus, large integer, UTF-8 strings, GC marking bit}

%\received{20 July 2023}
%\received[revised]{12 March 2009}
%\received[accepted]{5 June 2009}

%%
%% This command processes the author and affiliation and title
%% information and builds the first part of the formatted document.
\maketitle

\section{The revenge of address registers} 
\noindent
%% DCDC BSBS: c'est sympa comme retour en arrière, cette section, mais je pense que ce serait encore mieux si
%% on trouvait d'autres références et si on mettait quelques date, genre le permier MB par exemple, le premier GB, ??
%% Ou bien dans des subsection genre (chiffres bidons):
%%\subsection{1970--1980: 8 bit processor : max RAM size xx KB}
%%\subsection{1980--2000: 16 bit processor : max RAM size xx MB}
%% Ou bien un système plus light de commentaires entre parenthèse ??
%% Et aussi vérifier la cohérence avec une question que j'ai posé à CHAT GPT "Historically, what is the story of address registers size ?"
%% A méditer vu qu'on a de la place... 
%% Faudrait qu'on en cause au téléphone.
Several times in its history,
in the competition to increase the power of computers,
we've seen the emergence of tricks to first increase the size of the address bus before questioning the whole architecture.

For example, the mythical Z80 processor, with its 8-bit data bus and 16-bit addressing bus, should have been limited to 64 KB of RAM.
But, with its two-memory bank access principle, the Amstrad 6128 extends its memory to 128KB, doubling the capacity of its predecessor, the 464.

As part of the 8086 family with its 16-bit address register architecture, this processor has been equipped with a 20-bit address bus thanks to the addition of segment registers.
The address is made up of a pair of registers \{segment,offset\}.
The offset provides access to a contiguous 64KB range in memory.
As for the segment registers, they allow memory jumps in 16-byte steps.
Thus, the segment registers provide the 4 bits of high address required to reach the megabyte of RAM (1 MB).
The final address calculation is given by:
\begin{center}
  $Address = R_{\mathrm{segment}} \times 16 + R_{\mathrm{offset}}$
\end{center}

After a brief appearance of 24-bit processors with the 80286, came the 32-bit 80386 architecture, with a bus capable of addressing 4 GB of memory.
But again, in the last years of his reign, with a clever combination of segmentation and pagination,
the limit of this architecture has been pushed back to 36-bit addressing.

Then came 64-bit architecture, which for the first time in the history of computing has address registers that can go far beyond the memory capacities currently physically available.
The astronomical number of 16\,Eio ($2^{64} = 1.8e19$ bytes) that such a register could theoretically address is so far out of reach,
even in the distant future,
that designers preferred to truncate the logical address to 48 bits.
In this apparently arbitrary choice, the number of indirections needed to manage pagination must also be taken into account.
In fact, with 4 KB pages, we have the 12 least significant bits addressable contiguously,
then 4 indirection tables of 9 bits each must be consulted to reach a 48-bit physical address\footnote{
512 entrées par table d'indirection de 4KB: $2^9 \times 64$ bits = 4KB.
}.

So, quite surprisingly, this 48-bit logical address leaves a 16-bit high-order range unused for each address pointer location.
In addition, we can observe that memory allocators always allocate structures aligned at least with the machine word size.
So, if we consider a structure address, we also have the 3 least significant bits, which are always 0.
Interestingly, on a 64-bit architecture, all structure pointers have only 45 significant bits.
However, their use requires a few precautions before considering them as pointers.
It's worth noting that using a mask with a binary-and (\verb|&|) or a 3-bit binary shift to the left (\verb|<<|) are manipulations that have a marginal cost at runtime.
For each address, we therefore have 16 high-order bits at our disposal, and potentially 3 low-order bits that are free for other uses!

The thread running through this article is how best to use these insignificant bits in addresses to store precise information in a given context.
This practice is already widely used in the design of architectures and operating systems.
For example, in a pagination indirection table,
each page address is aligned on 4 KB,
so the 12 least significant bits are ignored when the MMU reads the page address.
These 12 unused bits contain other indicators, such as the right to write in this very memory page.
There are also other indicators defined and used solely by the operating system.

Further away from hardware and operating systems,
we also find this kind of approach at the software level, or more precisely at the language and compilation level in \cite{OOPSLA07}.
In Bonds et al (1992), {\it{}Tracking bad apples: Reporting the origin of null and undefined value errors},
the proposal is to internally replace the {\tt{}Null} value of a source program with deliberately invalid pointers to encode information.
Here, the information is used to trace the origin of the {\tt{}Null} in the event of an application crash.

Let's now look at how we can take advantage of all these findings concerning address format, starting with the management of numbers without overflow.

%% DCDC to BSBS: Désolé Ben, mais il va falloir relire et vérifier toute cette partie directement in english.
%% A cause des unités pour lesquelles il faut prendre la vesion english comme KB par exemple.
%% Et aussi le binary-and ... qui correspond pas au et-logique du français. (Le "logical and" c'est le et booléen... si j'ai bien pigé :-)  
%%

\section{Towards integers that don't overflow}
\noindent
\label{sec-num}
For the vast majority of programming languages, built-in integer types are limited to a certain size.
For example, \java{}'s {\tt{}int} type is limited to 32 bits.
A few rare cases of overflow are sometimes detected by the \java{} compiler only when the values are statically determinable.
In all cases, no overflow test is performed at runtime.
In this way, a positive value of type {\tt{}int} can be made negative by simply incrementing by 1.
This is often surprising, especially for novice programmers.

Another example from a more recent language is \rust{}, which notes this type {\tt{}i32} or {\tt{}u32} for the unsigned version.
In addition to the more appropriate type name, \rust{} in {\it{}debug} mode only, offers overflow control.
In {\it{}release} mode, overflow control is not performed.
As you might expect, \rust{}'s objective is to take full advantage of the processor's power.
There are even specialized functions to bypass overflow problems regardless of the compilation
mode\footnote{{\tt{}u32.wrapping\_add}, {\tt{}u64.wrapping\_add}, {\tt{}u64.wrapping\_add\_signed}, {\tt{}u64.wrapping\_sub}, etc.
}.
For large numbers without overflow, \rust{} also offers types such as {\tt{}BigInt} or {\tt{}BigUint}.

Historically, Smalltalk was the first language to natively integrate the concept of an integer that never overflows.
This choice is perfectly understandable in terms of comfort for programmers and for people unfamiliar with hardware constraints.
For Python, which is also interpreted, the choice is similar to Smalltalk:
no possiblle overflow and numbers that can grow in memory size as and when required.
In our opinion, there are two major drawbacks to this choice: slowness and the impossibility of easily taking hardware into account.

\subsection{Hardware limited types {\it{}and} flexible general type}
\noindent
In the Smalltalk language virtual machine, each 32-bit word representing an object is split into two parts\cite{SM80}.
The first part stores the information in 30 bits, the remaining two bits being the object's basic type.
We therefore have 4 internal object categories, one of which is reserved for a small 30-bit integer.
At runtime, if a calculation exceeds 30 bits, the small integer becomes a complex object with a pointer to model a larger integer via a real object which uses an array.
Smalltalk's object architecture was particularly well thought-out for its time, in the context of a 32-bit architecture and a pure, non-statically typed object environment.
However, we can only note a 30-bit address limit representing only 1 GB of accessible memory.
We've taken inspiration from this type of partitioning in the context of 64-bit architecture on a compiled language with static typing.

A good programming language needs to offer both programming comfort and full hardware speed.
The choice of the \rust{} language, which makes it possible to preserve types that can exactly match the characteristics of the hardware, must be maintained
(i.e. built-in types {\tt{}i8}, {\tt{}i32} and {\tt{}i64} for signed and built-in types {\tt{}u8}, {\tt{}u32} and {\tt{}u64} for unsigned).
Rather than resorting to general types specialized in handling numbers without overflow,
we propose a hybrid solution that allows,
according to variations in calculations,
to retain almost all the power of the small, limited numbers that exist natively.
In fact, depending on the evolution of calculations, for example, two very large numbers that are subtracted from each other can return to the 32-bit representation interval.
In such a case, it's interesting to get closer to the performance of native types for the result of this operation.

Unlike Smalltalk, which is completely and uniquely typed at runtime, we're working with a statically typed language.
For our proposal, it doesn't matter whether the typing is explicit or based on type inference.
Knowing that a given variable can only contain signed integers means that we can reuse the Smalltalk implementation idea with less variability in the entities represented.
So, unlike Smalltalk, and thanks to static typing, we have the whole machine word to best encode our integer.
In the following, we'll assume that the programming language offers a native type for handling signed integers, which we'll call {\typz}.
The idea is to always use a 64-bit machine word for any variable of type {\typz}.

Internally, and completely transparently to the user, there are three possible representations for the {\typz} type.
\mafigpdf{0.8}{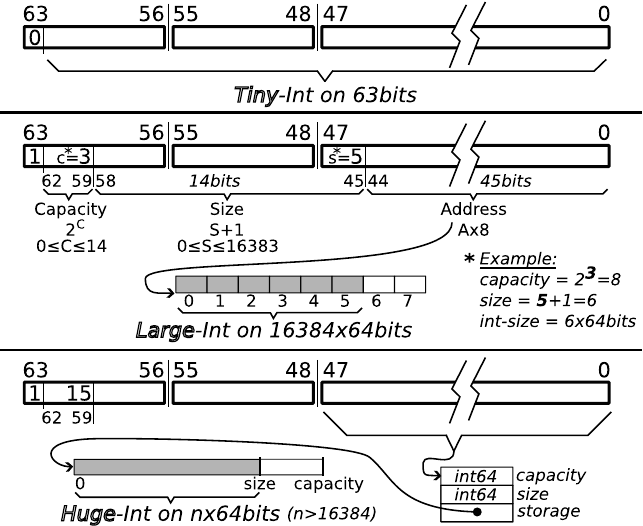}{The 3 integer encoding formats of the new {\typz} type.}
The aim is to make the most of 64-bit and get the best performance out of it.
The choice between these 3 representations is directly related to the size of the integer you need to model.
Figure \ref{fig-number} illustrates the following explanations of our 3 encoding formats for integers.

\subsection{{\tt{}Tiny} int: $n \le 63\:bits$ - see top of figure \ref{fig-number}}
\noindent
The first encoding format, {\tt{}Tiny}, is for a small integer that can be encoded on 63 bits, that is in two's complement
the range from $(-2^{62})$ to $(2^{62}-1)$.
This format is distinguished from the other two by the last bit 63 being set to 0.
Unsurprisingly, the use of the other 63 least significant bits stores our integer.
With a barely perceptible increase in computing time compared to using a 64-bit raw machine word, it enables fast management of small integers.
As with Smalltalk, when this capacity is exceeded, we dynamically migrate the integer to the second format we named {\tt{}Large}.

\subsection{{\tt{}Large} int: $64 bits \le n \le 2^{20}\,bits$ - middle of figure \ref{fig-number}}
\noindent
The second format for the {\typz} type, which we've named {\tt{}Large}, is the most complex and is particularly compact.
It can encode integers requiring a maximum of $16384 \times 64$ bits (i.e. an integer of 131\,072 bytes or 1 Mbits).
Thanks to its already particularly wide range, it's more than sufficient for most applications.
The 64-bit machine word representing the integer stores 3 pieces of information:
\begin{itemize}
\item
  The address of a contiguous memory area containing the integer using word of 64 bits.
\item
  The maximum capacity of this memory area.
  If necessary, the memory area is reallocated with a capacity twice that of the previous one\footnote{
  The well-known heuristic of doubling the capacity of a dynamic array is especially relevant to our use case.
  In this case, the multiplication of two $n$-bit integers uses $2\times n$ bits.
  }.
\item
  The size actually used in the memory area.
  In other words, the number of machine words needed to represent the integer in binary form.
\end{itemize}

The distribution of information within the 64 bits is as follows:
\begin{itemize}
\item[{\bf{}63}]
  Set to 1 to avoid being identified as a {\tt{}Tiny} encoding.
\item[{\bf{}59-62}]
  The capacity in power of 2.
  This 4-bit number can be used to encode capacities ranging from $2^0$ to $2^{14}$.
  Setting the 4 bits to 1 ($2^{15}$) is forbidden, and we reserve this value to identify the {\tt{}Huge} encoding format.
  We therefore have an array with a maximum capacity of 16384 64-bit cells.
\item[{\bf{}45-58}]
  This 14-bit range encodes the exact size actually used, from 1 to 16384.  
\item[{\bf{}0-44}]
  A 45-bit range representing the address of the corresponding allocated area.
  As this area is 64-bit aligned, a 3-bit left shift gives the exact 48-bit valid address.
\end{itemize}

\subsection{{\tt{}Huge} int: $2^{20} < n < $ whole memory - bottom fig. \ref{fig-number}}
\noindent
The last format for type {\typz} is shown at the bottom of figure \ref{fig-number} and is of a more standard design.
It is identified by the presence of 1 on all bits from 59 to 63.
The least significant part of the first 48 bits is an address to a standard object structure.
The corresponding object contains a {\tt{}capacity}, {\tt{}size} and {\tt{}storage} field for the usual implementation of a dynamic size array containing our
{\tt{}Huge} integer\footnote{
This representation is named {\tt{}ArrayList} in the \java{} library.
In \cpp{} this data structure is also known as {\tt{}std::vector}.
}.

\subsection{Integer decoding pseudo-code}
\noindent
Algorithm \ref{tiny-large-huge} illustrates in pseudo-code the decoding of different formats of the type {\typz}.
\begin{algorithm}
\caption{Reading an integer format {\tt{}Tiny}, {\tt{}Large} or {\tt{}Huge}}
\label{tiny-large-huge}
\begin{algorithmic} 
  \IF{$(W \ge 0)$}
  \STATE {\tt{}// Format} {\bf{}TINY}
  \STATE $INT \gets ((((signed 64)W)\,<<\,1)\,>>\,1)$
  \ELSE
  \STATE $cap \gets (W \gg 58)\:\&\:\mathrm{F}h$
  \IF{$(cap = 1111b)$}
  \STATE {\tt{}// Format} {\bf{}HUGE}
  \STATE $a \gets W\:\&\:\mathrm{FFFF\,FFFF\,FFFF}h$
  \STATE $buf \gets a.\mathrm{storage}$
  \STATE $siz \gets a.\mathrm{size}$
  \STATE $cap \gets a.\mathrm{capacity}$
  \ELSE
  \STATE {\tt{}// Format} {\bf{}LARGE}
  \STATE $buf \gets (W\:\&\:\mathrm{1FFF\,FFFF\,FFFF}h) \ll 3$  
  \STATE $siz \gets ((W \gg 44)\:\&\:\mathrm{3FFF}h) + 1$
  \STATE $cap \gets 1 \ll cap$  
  \ENDIF
  \STATE $INT \gets \textbf{tab}(buf, siz, cap)$
  \ENDIF
\end{algorithmic}
\end{algorithm}
In this algorithm $W$ is the machine word representing the integer encoded in one of the 3 formats, and $INT$ represents bit access to the integer.
Note that when using a small integer of less than 64 bits, the extra cost compared to the standard 64-bit basic type is just sign detection (or the position of bit 63) and a jump.

Note also that our representation is deliberately canonical.
Depending on its size in number of bits, a given number has to be represented in just one of the three categories, {\tt{}Tiny}, {\tt{}Large} or {\tt{}Huge}.
This makes it easy to compare numbers with each other.
In particular, the comparison between two numbers {\tt{}Tiny} is made with the usual machine instruction for comparing two memory words.

\section{String indexing in \utf{8} format}
\noindent
\label{sec-utf}
The use of the \utf{8} format for strings has become widespread.
The main advantage is its compatibility with the standard {\ascii} format.
It also offers compact encoding of {\unicode} characters strings, with variable character sizes.
Its encoding allows a string to be read in both directions, which means that many older, highly efficient string processing algorithms remain applicable.
The penalizing management of \utf{16} in native as in \java{} is no longer desirable.
Today, the development of a new programming language clearly suggests native management of \utf{8}.

Compared with {\ascii} and despite its advantages, handling \utf{8} remains difficult and rather unnatural due to the variable character size.
The conversion of an \utf{8} character into {\unicode} contained in an array of bytes is performed by the standard algorithm \ref{utf8-to-unicode}.
\begin{algorithm}
\caption{Standard \utf{8} to {\unicode} conversion.}
\label{utf8-to-unicode}
\begin{algorithmic}
  \IF{$ptr[0] < 80h$}
  \STATE $ucode \gets ptr[0]$
  \ELSIF{$(ptr[0] \:\&\: \mathrm{E0}h) = C0h$ {\bf{}and}\\
    \hspace{1em}$(ptr[1] \:\&\: \mathrm{C0}h) = 80h$}
  \STATE $ucode \gets ((ptr[0] \:\&\: \mathrm{1F}h) \ll 6) \:|\: (ptr[1] \:\&\: \mathrm{3F}h)$
  \ELSIF{$(ptr[0] \:\&\: \mathrm{F0}h) = \mathrm{E0}h$ {\bf{}and}\\
    \hspace{1em}$(ptr[1] \:\&\: \mathrm{C0}h) = 80h$ {\bf{}and}\\
    \hspace{1em}$(ptr[2] \:\&\: \mathrm{C0}h) = 80h$}
  \STATE $ucode \gets ((ptr[0] \:\&\: \mathrm{0F}h) \ll 12) \:|\:$
  \STATE \hspace{1em}$((ptr[1] \:\&\: \mathrm{3F}h) \ll 6) \:|\:(ptr[2] \:\&\: \mathrm{3F}h)$
  \ELSIF{$(ptr[0] \:\&\: \mathrm{F8}h) = \mathrm{F0}h$ {\bf{}and} \\
    \hspace{1em}$(ptr[1] \:\&\: \mathrm{C0}h) = 80h$ {\bf{}and} \\
    \hspace{1em}$(ptr[2] \:\&\: \mathrm{C0}h) = 80h$ {\bf{}and} \\
    \hspace{1em}$(ptr[3] \:\&\: \mathrm{C0}h) = 80h$}
  \STATE $ucode \gets ((ptr[0] \:\&\: 07h) \ll 18) \:|\:$\\
  \STATE $((ptr[1] \:\&\: \mathrm{3F}h) \ll 12) \:|\:((ptr[2] \:\&\: \mathrm{3F}h) \ll 6) \:|\:$
  \STATE $(ptr[3] \:\&\: \mathrm{3F}h)$
  \ELSE
  \STATE \textbf{print} "Invalid UTF-8 sequence"
  \ENDIF
  \STATE {\bf{}Output} $ucode$
\end{algorithmic}
\end{algorithm}
Processing and calculating a string index is much trickier.
A number of libraries are currently available
({\tt{}ICU}, {\tt{}UTF8-CPP}, {\tt{}libunistring}, {\tt{}libiconv},\ldots)
and are essentially based on two approaches to the treatment of \utf{8}.

The first approach is to provide an interface for converting from \utf{8} to {\unicode} and vice versa.
Thus, the string translated into {\unicode} is made up of 32-bit words, i.e. one 32-bit word for each character.
With the exception of very large strings, processing becomes as easy as in the days of the {\ascii} format.
The disadvantages of this technique are, of course, the translation from one format to another, and the fact that in the vast majority of cases, memory representation is quite costly.

The second approach to managing \utf{8} strings is to offer the user high-level functions that don't require indexes.
Its principle is based on the use of regular expressions similar to those found in {\tt{}grep}, {\tt{}sed} ou {\tt{}awk}.
It's a good choice in many situations.
However, these are complex mechanisms and their intensive use can quickly become a drag in terms of performance.
Furthermore, the programmer can quickly become frustrated at not having direct and easy access to the internal structure of the string via character-by-character indexing.

We propose a different approach to virtualize the notion of index.
Our technique is necessarily slightly more expensive than a simple {\ascii} string, but no more expensive than a standard implementation of \utf{8} string management.
The programmer can use indexes to consider a character jump as always being a single logical step.
The corresponding physical step in memory is variable, but remains transparent to the user.

Our library is based on a single object type {\tt{}StrIdx} modeled by a single 64-bit word.
The idea is to represent the logical index and the physical index in the same word, split into two parts of 32 bits each.
In the example shown in figure \ref{fig-utf8}, the logical index of the first letter '{\tt{}n}' is $4$ and its physical index is $5$,
due to the presence of the letter 'ç' before it.  
\mafigpdf{0.8}{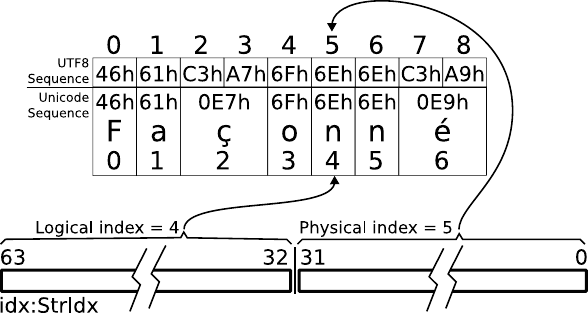}{Type {\tt{}StrIdx} includes both logical and physical index encoding. All in 64-bit memory.}
With this representation, in the worst case, the ratio between a \utf{8} character and its memory occupancy is one to 4.
With an arbitrary division of 32 bits for the physical part and 32 bits for the logical part, we can process contiguous strings of 1 to 4 billion characters.

So, with a single 64-bit word, both the logical index and the corresponding physical index are stored.
Note that if {\tt{}idx} is of type {\tt{}StrIdx}, the expression ($\text{idx} \gg 32$) gives access to the logical index.
The physical index is obtained with the ($\text{idx}\:\&\:\text{FFFF\,FFFF}h$) expression.
\begin{algorithm}
\caption{Move $\text{idx}$ of type {\tt{}StrIdx} one step forward.}
\label{next-character}
\begin{algorithmic}    
  \STATE $\text{code} \gets \text{str.at((int32)idx)}$
  \STATE $\text{idx} \gets \text{idx} + 1\,0000\,0001h$        
  \IF{$(\text{code}\:\&\:80h)$}
  \REPEAT    
  \STATE $\text{code} \gets \text{code} \ll 1$
  \STATE $\text{idx} \gets \text{idx} + 1$
  \UNTIL{$(\text{code}\:\&\:\text{40}h = 0)$}              
  \ENDIF
  \STATE \textbf{Output} $\text{idx}$
\end{algorithmic}
\end{algorithm}
Algorithm \ref{next-character} calculates the next {\tt{}StrIdx} index from a given {\tt{}StrIdx} index.
\begin{algorithm}
\caption{Move $\text{idx}$ of type {\tt{}StrIdx} one step backward.}
\label{previous-character}
\begin{algorithmic}        
  \STATE $\text{idx} \gets \text{idx} - 1\,0000\,0001h$        
  \WHILE{$(\text{str.at((int32)idx)}\:\&\:80h)$}
  \STATE $\text{idx} \gets \text{idx} - 1$
  \ENDWHILE
  \STATE \textbf{Output} $\text{idx}$
\end{algorithmic}
\end{algorithm}
Moving back one character is also easy to calculate (see algorithm \ref{previous-character}).
It's worth noting that there's no extra cost involved in managing our index compared to what's usually done when handling the \utf{8} format.
With advanced languages capable of defining binary operators, it is possible to facilitate the use of the two preceding algorithms.
With the definition of a '+' and '-' operator taking as receiver an index of type {\tt{}StrIdx} and as parameter an integer,
we can move the index by simple arithmetic operation by any number of characters, regardless of their size.
\begin{algorithm}
\caption{Move forward and compute unicode skipped.}
\label{unicode-forward}
\begin{algorithmic}    
  \STATE $\text{ucode} \gets \text{str.at((int32)idx)}$
  \STATE $\text{idx} \gets \text{idx} + 1\,0000\,0001h$        
  \IF{$(\text{ucode}\:\&\:80h)$}
  \STATE $\text{msk} \gets 40h$          
  \REPEAT
  \STATE $\text{ucode} \gets (\text{ucode} \ll 6) \mid (\text{str.at((int32)idx)}\:\&\:3\mathrm{F}h)$
  \STATE $\text{msk} \gets \text{msk} \ll 5$
  \STATE $\text{idx} \gets \text{idx} + 1$
  \UNTIL{$(\text{ucode}\:\&\:\text{msk} = 0)$}          
  \STATE $\text{ucode} \gets \text{ucode}\:\&\:(\text{msk} - 1)$
  \ENDIF        
  \STATE \textbf{Output} $\text{ucode},\text{idx}$
\end{algorithmic}
\end{algorithm}
Finally, we propose Algorithm \ref{unicode-forward}, which is an optimized version of Algorithm \ref{utf8-to-unicode}, enabling a very frequent operation:
calculate the current unicode and advance the index to the next character.

\section{The marking bit for the garbage collector}
\noindent
\label{sec-gc}
% Dans CLONE, mettre ! Self:SELF :SELF <- clone
Among programming languages, the mark-and-sweep algorithm is still the most widely used for automatic memory management \cite{Wil92}, \cite{gcJones1996}, \cite{gcJones2011}.
Without going into detail, this algorithm stops the program in progress, and performs two distinct phases:
\begin{enumerate}
\item
  Starting with the objects directly accessible at time $t$ (stack, global variables) during program execution,
  the marking phase consists of recursively traversing these objects and marking them with a flag as being a live object.
\item
  In a second step, a linear memory run frees unmarked objects and cancels the marking of other objects.
\end{enumerate}
After these two phases, the program can resume execution until the garbage collector is triggered again.
Many variants of this algorithm exist, in particular to avoid program interruption,
but the use of a marking bit is still required for its implementation.

The question for any language designer intending to implement this type of garbage collector is where to store this marking bit
without polluting the structure of each allocated object.
In many cases, this is not such a problem: adding an attribute within the object structure is easy to do.
The waste-conscious designer may be sensitive to the systematic allocation of 64 bits for each object, when only one bit is necessary.
But that's not the only reason to dwell on this issue.
In low-level programming, it's not uncommon for a critical section of an application to be made up of objects with a frozen, hardware-imposed structure.
Low-level programming is not incompatible with high-level object-oriented languages like {\tt{}Lisaac} \cite{sonntag:inria-00100788}, \cite{Lisaac}.
If this is the case, there's no question of shamelessly adding a field to the structure for memory management.

In \cite{colnet:inria-00098708},\cite{SPAE14} we see that it is possible to specialize the marking code according to the type of the object.
This work focuses on how to recursively traverse the object graph.
Here, our aim is to perform a specialization to change the value of this marking bit.
With this specialization, we can find a specific place for each category or type of object.

Our idea is to find one or more unused bits in each structure representing a given type.
In the following, we present several places where you can easily find this space.

\subsection{Use an attribute that references another object}
\noindent
It's very common for the structure of an object to have attributes that are references to other objects.
We therefore have the 16 most significant bits which are not used by the corresponding pointer.
So, for example, the most significant bit can be used as the marking bit for garbage collection.
\begin{algorithm}
\caption{Setting the mark bit during the mark phase.}
\label{gc-markbit}
\begin{algorithmic}    
  \STATE ${\bf{}mark\_}{\it{}type\_name}(\mathrm{obj}) \gets$
  \STATE $\mathrm{ptr} \gets \mathrm{obj}.{\it{}attribute}$
  \IF{$(\mathrm{ptr}\:\&\:8000\,0000\,0000\,0000h = 0)$}  
  \STATE $\mathrm{obj}.{\it{}attribute} \gets \mathrm{ptr} \mid 8000\,0000\,0000\,0000h$
  \STATE ${\bf{}mark\_}{\it{}attribute.type\_name}(\mathrm{ptr})$
  \STATE ${\it{}... Handling\:of\:other\:attributes ...}$
  \ENDIF        
\end{algorithmic}
\end{algorithm}
Algorithm \ref{gc-markbit} shows how and where you can specialize the marking routine to further specialize the code in \cite{colnet:inria-00098708}.
To avoid complicating the code unnecessarily, we've assumed that the attribute type is monomorphic.
\begin{algorithm}
\caption{Clearing the mark bit during sweep phase.}
\label{mark-bit-sweep}
\begin{algorithmic}    
  \STATE ${\bf{}sweep\_}{\it{}type\_name}(\mathrm{obj}) \gets$  
  \IF{$(\mathrm{obj}.{\it{}attribute}\:\&\:8000\,0000\,0000\,0000h)$}
  \STATE $\mathrm{obj}.{\it{}attribute} \gets \mathrm{obj}.{\it{}attribute}\:\&\,\mathrm{7FFF\,FFFF\,FFFF\,FFFF}h$
  \ELSE
  \STATE ${\bf{}free}(\mathrm{obj})$  
  \ENDIF        
\end{algorithmic}
\end{algorithm}
Of course, the most significant bit of the pointer in question must be cleared during the sweep phase (see algorithm \ref{mark-bit-sweep}).

\subsection{Use the dynamic dispatch identification field}
\noindent
If an object's type is subject to polymorphic calls, there is a hidden attribute at the beginning of its structure to identify its type.
This field is necessary to handle dynamic dispatch in polymorphic calls.
Without going into the details described in \cite{bdboopsla97},
this identifier is a simple integer with a maximum size directly related to the number of polymorphic types to be discriminated.
In the optimization context described in \cite{SPAE12}, even if we globalize this identifier to the number of types possible in a given application,
this number very rarely exceeds 8 bits.
%%Bien entendu, même si c'est un pointeur vers une table d'indirection, il reste également de la place.
\begin{algorithm}
\caption{The marking bit inside the dynamic dispatch information field.}
\label{id-markbit}
\begin{algorithmic}
  \STATE $\mathrm{ptr} \gets \mathrm{obj}.{\it{}attribute}$
  \STATE $\mathrm{id} \gets \mathrm{ptr.id}$
  \IF{$(\mathrm{id}\:\&\:8000\,0000\,0000\,0000h = 0)$}
  \STATE $\mathrm{ptr.id} \gets \mathrm{id} \mid 8000\,0000\,0000\,0000h$
  \STATE ${\bf{}dispatch\_and\_mark}(\mathrm{id,ptr})$
  \STATE ${\it{}... Handling\:of\:other\: attributes ...}$
  \ENDIF        
\end{algorithmic}
\end{algorithm}
We therefore have all the space we need to install the marking bit at this point, always using specialized methods.
However, it is important to note that it is necessary to take into account this bit potentially marked during the first phase of the recursive path (see algorithm \ref{id-markbit}).

\subsection{Use the re-alignment space in the data structure}
\noindent
Finally, if none of the above is possible, we can consider re-alignment zones.
An attribute with a size of less than 64 bits will be compile-aligned to a 64-bit word.
By adding a one-byte attribute at this precise location, the overall size of the structure is not increased at all.
This byte can then be used as a marker bit.

In the example below, the two structures {\tt{}obj\_A} and {\tt{}obj\_B} are both 16 bytes long.
Figure \ref{fig-struct} illustrates the alignment and memory representation of the {\tt{}obj\_B} structure.

\noindent
%\begin{tabular*}{0.48\textwidth}{@{\extracolsep{\fill}}l|l}
\begin{tabular*}{0.48\textwidth}{l l l l}
  \hline
&  {\it{}struct} {\tt{}obj\_A} \{       & & {\it{}struct} {\tt{}obj\_B} \{ \\
&  \hspace{1em}{\it{}int} {\tt{}att0};  & & \hspace{1em}{\it{}int} {\tt{}att0}; \\
&  \hspace{1em}{\it{}long} {\tt{}att1}; & & \hspace{1em}{\it{}char} {\bf{}flag}; \\
&  \}                                   & & \hspace{1em}{\it{}long} {\tt{}att1}; \\
\hspace{4em}&  {}                       & \hspace{5em} & \}\\
  \hline     
\end{tabular*}
\mafigpdf{0.95}{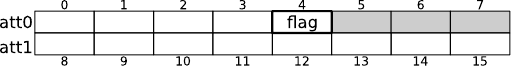}{Memory representation of the  {\tt{}obj\_B} structure.}

As we've shown, it's very often possible to find enough unused space.
These free slots can be perfectly used for more complex variants of the {\it{}mark-and-sweep} requiring more than one marking bit.
This technique is particularly well suited to garbage collectors whose code is specialized at compile time.

This free space can also be used for the implementation of monitors.
Like \java{}'s {\tt{}synchronized} blocks, for example.
In fact, it's usually easy to find 8 or even 16 bits within an object.
This value can then be used as an index in a global table of pointers to the corresponding threads.

\section{Conclusion}
\noindent
We've known since its emergence that 64-bit architectures tend to make executables and data structures mechanically larger than 32-bit architectures.
Many bits are indeed wasted, often due to allocation alignments.
But this wastage is also present in the fact that a pointer is limited to 48 bits, whereas it is represented in memory by a 64-bit word.

As part of the library implementation of a new language, we present here three very interesting uses for these wasted bits.

First, we give different physical representations of a multi-precision integer taking advantage of 64-bit architectures to gain performance in section \ref{sec-num}.

In section \ref{sec-utf}, we look at a new approach to \utf{8} string manipulation by index virtualization using a 64-bit word.

Finally, section \ref{sec-gc} shows how to optimize storage of the marking bit required in the implementation of the {\it{}mark \& sweep} garbage collector algorithm.

%% DCDC en plus, ya déjà un max de gachis si on utilise malloc pour un petit truc. On paume 2x64 bits vu qu'il faut réalligner et être sur de pouvoir indiquer la taille.

\bibliographystyle{unsrt}
\bibliography{biblio}
%\nocite{colnet:hal-01116288}

\end{document}